\documentclass[conference]{IEEEtran}
\IEEEoverridecommandlockouts

\usepackage{cite}
\usepackage{amsmath,amssymb,amsfonts}

\usepackage{graphicx}
\usepackage{textcomp}
\usepackage{url}
\usepackage{xcolor}
\def\BibTeX{{\rm B\kern-.05em{\sc i\kern-.025em b}\kern-.08em
    T\kern-.1667em\lower.7ex\hbox{E}\kern-.125emX}}
\begin{document}

\title{ForestBack: Breadcrumb-Based Pedestrian Dead Reckoning for Infrastructure-Free Return Navigation}

\author{
\IEEEauthorblockN{1\textsuperscript{st} Aueaphum Aueawatthanaphisut}
\IEEEauthorblockA{
\textit{School of Information, Computer, and Communication Technology} \\
\textit{Sirindhorn International Institute of Technology, Thammasat University} \\
Pathum Thani, Thailand \\
aueawatth.aue@gmail.com
}

\and

\IEEEauthorblockN{2\textsuperscript{nd} Chanakan Chaipan}
\IEEEauthorblockA{
\textit{English Program} \\
\textit{Samsenwittayalai School} \\
Bangkok, Thailand \\
chanakanc5999@gmail.com
}
}

\maketitle

\begin{abstract}
Reliable return navigation remains an important challenge in GPS-denied environments where external positioning infrastructure may be unavailable or unreliable. This paper presents \textit{ForestBack}, an infrastructure-free pedestrian return navigation framework based on breadcrumb-based pedestrian dead reckoning (PDR). The proposed system records a user's walking route as a sequence of reversible breadcrumb nodes and generates reverse-path guidance without requiring GPS, Wi-Fi, Bluetooth beacons, or pre-installed infrastructure. The framework integrates acceleration-based step detection, adaptive step-length estimation, magnetometer-assisted heading estimation, barometric-altitude correction, and bidirectional breadcrumb path reconstruction. To evaluate the proposed approach under a controlled navigation scenario, an indoor obstacle-avoidance test route was designed using five checkpoints from A to E, where the user was required to navigate around a central obstacle. A high-resolution experimental dataset consisting of 36 walking trials and 42,474 time-series samples was used for evaluation. The dataset includes IMU signals, magnetometer readings, barometric variables, turn-event labels, ground-truth trajectories, baseline PDR outputs, proposed ForestBack outputs, and power-related measurements, and is publicly available in the ForestBack-Dataset repository~\cite{b11}. Experimental analysis shows that the proposed ForestBack method reduced the mean RMSE from 1.129 m to 0.965 m compared with traditional PDR, corresponding to an average improvement of 15.76\%. The mean final-position error was also reduced from 1.781 m to 1.388 m, while turn-event detection consistency reached approximately 99.90\%. These results indicate that ForestBack can improve trajectory reconstruction and route-preserving return guidance in obstacle-avoidance scenarios. The released dataset and analysis notebook support reproducibility and future benchmarking of infrastructure-free PDR-based return navigation systems.
\end{abstract}


\section{Introduction}

\subsection{Background of GPS-Denied Navigation}

Global Navigation Satellite Systems (GNSS), such as the Global Positioning System (GPS), are widely used for outdoor positioning and navigation. In open-sky environments, GNSS receivers can estimate position accurately because satellite signals can be received with minimal obstruction. However, GNSS performance degrades significantly in environments where satellite signals are blocked, attenuated, or affected by multipath propagation. Examples of such GPS-denied environments include dense forests, indoor spaces, underground areas, caves, and urban canyons. In these scenarios, reliable positioning becomes difficult because external satellite-based references may be unavailable or unstable~\cite{b1}.

To address this limitation, alternative navigation methods have been investigated for applications that require positioning without continuous GNSS availability. Such applications include indoor navigation, underground exploration, search-and-rescue operations, and infrastructure-free return guidance. In these cases, localization must be performed using onboard sensors or local environmental observations instead of relying on external positioning infrastructure~\cite{b10}.

Pedestrian Dead Reckoning (PDR) is one of the most common approaches for navigation in GPS-denied environments. PDR estimates a pedestrian's relative movement by detecting steps, estimating step length, and determining walking direction using sensors such as accelerometers, gyroscopes, and magnetometers. Instead of estimating absolute position from external signals, PDR updates the user's position relative to a known starting point. This makes PDR suitable for lightweight and infrastructure-free navigation systems, especially when the objective is to reconstruct a walking path or guide a user back to a starting location~\cite{b1}.

\subsection{Pedestrian Dead Reckoning}

Pedestrian Dead Reckoning estimates a user's position by incrementally updating the previous position according to detected walking motion. A typical PDR system consists of three main components: step detection, step-length estimation, and heading estimation. Step detection identifies when a user takes a step, step-length estimation determines the distance traveled during each step, and heading estimation determines the walking direction~\cite{b2},~\cite{b4}.

Step detection is commonly performed using accelerometer measurements. Since human walking produces periodic acceleration patterns, peaks or periodic components in the acceleration magnitude can be used to identify step events. Once a step is detected, the system estimates the step length using either a constant model, an empirical model, or an adaptive model based on walking dynamics~\cite{b2}. Accurate step-length estimation is important because even small distance errors can accumulate over time and cause significant trajectory deviation.

Heading estimation is another critical component of PDR. It is usually obtained from gyroscope integration, magnetometer-based compass readings, or sensor fusion methods that combine multiple sensor measurements. Since each PDR update depends on both the estimated step length and the heading angle, heading error can strongly affect the reconstructed trajectory. Small heading errors may lead to large position errors when accumulated over many steps~\cite{b4}.

\subsection{Limitations of Existing PDR Systems}

Although PDR is suitable for infrastructure-free pedestrian navigation, several limitations remain. The most important limitation is accumulated drift. Since the estimated position is continuously updated from previous estimates, errors in step detection, step length, or heading can accumulate over time. As a result, the reconstructed trajectory may gradually diverge from the actual walking path~\cite{b4},~\cite{b7}.

Sensor noise and bias also affect PDR performance. Low-cost inertial measurement units (IMUs), which are commonly used in wearable devices and smartphones, are affected by accelerometer noise, gyroscope bias, and measurement uncertainty. These errors can degrade step detection, heading estimation, and trajectory reconstruction, particularly during long-duration walking or routes containing multiple turns~\cite{b7}.

Magnetic disturbance is another major challenge. Magnetometers are often used to correct heading drift, but their measurements can be affected by nearby metal objects, electronic devices, building structures, and other magnetic anomalies. Such disturbances may produce incorrect heading estimates and lead to trajectory distortion~\cite{b5}. Therefore, robust heading estimation remains an important issue for PDR-based navigation.

In addition, many existing PDR systems are designed mainly for indoor positioning or smartphone-based localization. Some approaches rely on Wi-Fi, Bluetooth, magnetic maps, or pre-installed infrastructure to correct accumulated error. Although these methods can improve positioning accuracy, they may not be suitable for infrastructure-free return navigation, where no external references are available~\cite{b10}.

\subsection{Research Gap}

Existing PDR studies have made significant progress in indoor localization, smartphone-based positioning, and sensor fusion. However, several gaps remain for infrastructure-free return navigation. First, many systems focus on estimating the user's current position rather than preserving a reversible path for return guidance. Second, infrastructure-assisted correction methods such as Wi-Fi, Bluetooth, or map matching may not be available in GPS-denied or remote environments. Third, many solutions require additional computation or environmental preparation, which may limit their use on low-power embedded platforms.

For return navigation, the system does not necessarily need to estimate an absolute global position. Instead, it must preserve the relative structure of the user's walking route and provide reliable reverse-path guidance. This motivates the development of a breadcrumb-based PDR framework that records the walking route as a sequence of relative movement nodes and converts the recorded path into return guidance commands.

\subsection{Contributions of This Work}

To address the limitations of existing PDR systems for infrastructure-free return navigation, this paper proposes \textit{ForestBack}, a breadcrumb-based PDR framework designed to record relative walking paths and generate reverse-path guidance without external positioning infrastructure. The system uses pedestrian dead reckoning together with IMU sensing, magnetometer-assisted heading estimation, and barometric-altitude correction to estimate user movement and support return navigation~\cite{b2},~\cite{b3}.

The main contributions of this work are summarized as follows:

\begin{itemize}
    \item A breadcrumb-based PDR framework is proposed for infrastructure-free return navigation.
    \item An adaptive step-length estimation model is introduced to improve relative distance estimation under varying walking dynamics.
    \item Magnetometer-assisted heading estimation and barometric-altitude information are incorporated to support trajectory reconstruction.
    \item A bidirectional breadcrumb return mechanism is developed to convert the recorded walking path into reverse-path guidance commands.
    \item An experimental obstacle-avoidance dataset and analysis notebook are released to support reproducibility and future benchmarking.
\end{itemize}

\section{Related Work}

\subsection{Inertial Navigation Systems for Pedestrians}

Inertial Navigation Systems (INS) are widely used for pedestrian navigation when external positioning signals are unavailable. An INS estimates motion using an inertial measurement unit, typically consisting of accelerometers and gyroscopes. Accelerometers measure linear acceleration, while gyroscopes measure angular velocity. By integrating these measurements over time, pedestrian motion and orientation can be estimated. However, pure inertial navigation is highly sensitive to sensor noise and bias. Small errors in acceleration or angular velocity may accumulate rapidly, resulting in significant position drift~\cite{b7}.

To reduce inertial drift, pedestrian navigation systems often combine INS with PDR. Instead of continuously integrating acceleration to estimate position, PDR updates the pedestrian position at detected step events. This step-based update structure reduces some of the instability associated with pure inertial integration because human walking motion has periodic and constrained patterns.

Montorsi \textit{et al.} developed a pedestrian inertial navigation system using a low-cost MEMS IMU and filtering-based error correction~\cite{b7}. Their work demonstrated that filtering methods and walking constraints can improve pedestrian navigation accuracy. Zero Velocity Update (ZUPT) is another commonly used correction technique, particularly for foot-mounted systems. During the stance phase of walking, the foot is assumed to have zero velocity, allowing accumulated velocity errors to be corrected. Although these methods reduce drift, they may require specific sensor placement or additional assumptions about walking motion.

\subsection{Smartphone-Based Pedestrian Dead Reckoning}

Smartphones are frequently used as platforms for PDR research because they contain built-in accelerometers, gyroscopes, and magnetometers. These sensors allow smartphones to estimate step events, walking direction, and relative displacement without external positioning infrastructure. As a result, smartphone-based PDR has been widely studied for indoor localization and GPS-denied navigation~\cite{b2},~\cite{b3}.

A typical smartphone-based PDR system includes step detection, step-length estimation, and heading estimation. Step detection identifies walking events from acceleration patterns. Step-length estimation determines the distance traveled during each detected step, while heading estimation determines the walking direction. The estimated position is then updated step by step. However, the accuracy of smartphone-based PDR is affected by device placement, sensor noise, user walking variability, and magnetic interference~\cite{b4}.

Several studies have proposed robust step detection and heading estimation methods to improve smartphone-based PDR accuracy. Kuang \textit{et al.} investigated robust PDR using MEMS-IMU sensors for smartphones and highlighted the importance of reliable step detection and heading estimation~\cite{b2}. Geng \textit{et al.} studied smartphone-based PDR for three-dimensional indoor positioning and demonstrated the potential of combining inertial and barometric information for vertical movement estimation~\cite{b3}. These studies show that smartphone sensors can support PDR-based navigation, but accumulated drift remains a major challenge.

\subsection{Heading Estimation and Magnetic Disturbance}

Heading estimation is a critical factor in PDR because position updates depend directly on the walking direction. Gyroscopes provide short-term rotational information, but their estimates drift over time due to bias. Magnetometers can provide an absolute heading reference relative to the Earth's magnetic field, but they are sensitive to environmental magnetic disturbance~\cite{b4},~\cite{b5}.

Magnetic disturbance commonly occurs near metal structures, electronic devices, and indoor infrastructure. Such disturbance can corrupt magnetometer readings and produce incorrect heading estimates. Since PDR continuously updates the trajectory using heading information, even a small heading error can lead to a large position error after many steps. Therefore, heading estimation under magnetic interference has been a major research topic in PDR systems.

Li \textit{et al.} investigated heading estimation under magnetic interference and multiple smartphone postures, showing that robust heading estimation is essential for reliable PDR performance~\cite{b4}. Zhu \textit{et al.} further explored the use of deep learning to improve indoor PDR under magnetic interference~\cite{b5}. Although these approaches can improve heading robustness, they may require additional training data, computational resources, or device-specific calibration.

\subsection{SLAM and Sensor Fusion Approaches}

Sensor fusion methods have been widely used to improve pedestrian navigation accuracy by combining information from multiple sensors. Accelerometer, gyroscope, magnetometer, and barometer measurements can provide complementary information about pedestrian motion, orientation, and altitude. Filtering methods such as Kalman filtering and particle filtering are commonly used to combine these sensor measurements and reduce uncertainty~\cite{b7},~\cite{b10}.

Simultaneous Localization and Mapping (SLAM) has also been applied to pedestrian navigation. In SLAM-based systems, the user's position is estimated while a representation of the environment is simultaneously constructed. Some approaches use particle filters to maintain multiple possible position hypotheses and update them according to sensor measurements and environmental constraints~\cite{b8}. Other approaches combine PDR with magnetic field matching to correct accumulated position error using magnetic maps~\cite{b6}.

Although SLAM and map-matching approaches can improve positioning accuracy, they often require environmental information, prior maps, repeated observations, or higher computational resources. These requirements may limit their use in lightweight infrastructure-free return navigation systems. In contrast, breadcrumb-based PDR focuses on preserving the relative walking route and generating return guidance without requiring a pre-existing map or external infrastructure.

\subsection{Summary of Research Gap}

Previous studies have demonstrated that PDR, INS, sensor fusion, magnetic matching, and SLAM can support pedestrian navigation in GPS-denied environments. However, many systems are designed primarily for indoor localization, smartphone positioning, or map-assisted correction. These systems may not directly address the problem of infrastructure-free return navigation, where the objective is to record a walking path and guide the user back to the starting point without external references.

Therefore, a lightweight PDR framework that records relative pedestrian movement as reversible breadcrumb nodes remains useful. Such a framework can support return guidance in GPS-denied and obstacle-constrained environments while avoiding reliance on Wi-Fi, Bluetooth beacons, pre-installed maps, or other external infrastructure. This paper addresses this gap by proposing \textit{ForestBack}, a breadcrumb-based PDR framework for infrastructure-free return navigation.

\section{Methodology}

This section presents the methodology of the proposed \textit{ForestBack} system, an infrastructure-free breadcrumb-based return navigation framework using pedestrian dead reckoning (PDR), IMU sensing, magnetometer-assisted heading estimation, and barometric-altitude correction.

\begin{figure*}[h]
    \centering
    \includegraphics[width=1\linewidth]{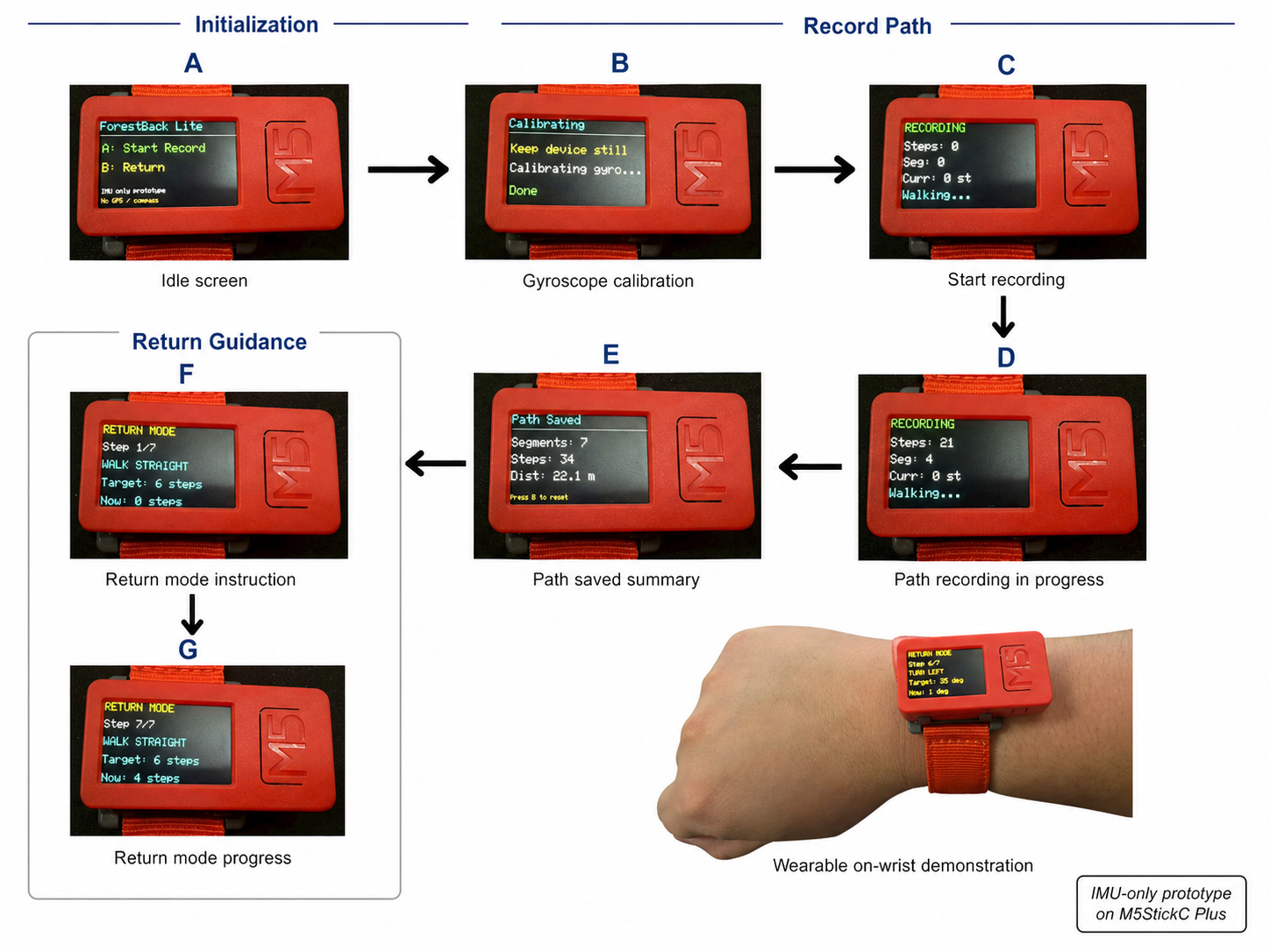}
    \caption{Prototype workflow of the proposed ForestBack wearable navigation system implemented on an M5StickC Plus. The system starts from the idle screen and gyroscope calibration stage, then records the user's walking path as breadcrumb segments during forward navigation. After the path is saved, the device switches to return guidance mode, where step-by-step instructions such as walking straight or turning are displayed to guide the user back along the recorded route.}
\label{fig:forestback_prototype_workflow}
\end{figure*}

\subsection{System Overview of ForestBack}

The proposed system in Fig.1 aims to provide infrastructure-free return navigation capability in GPS-denied and obstacle-constrained environments. The architecture consists of multiple onboard sensors integrated with a low-power microcontroller.

\begin{itemize}
\item Inertial Measurement Unit (IMU)
\item Magnetometer
\item Barometer
\item Microcontroller
\end{itemize}

The system estimates pedestrian motion through inertial sensing and reconstructs the trajectory using a pedestrian dead reckoning algorithm.


\subsection{Experimental Protocol and Ground Truth Definition}

The indoor obstacle-avoidance evaluation was conducted using a predefined checkpoint-based route consisting of five reference points, denoted as A, B, C, D, and E. Checkpoint A was defined as the starting point, while checkpoint E was defined as the final destination. The route was designed to include diagonal, horizontal, and vertical walking segments around a central obstacle so that step detection, heading transition, and accumulated drift could be evaluated under a controlled route-following condition.

The ground-truth trajectory was defined from the measured checkpoint layout of the indoor test map. Each checkpoint coordinate was assigned according to the physical route geometry, and the ground-truth path was constructed by linearly connecting the ordered checkpoint sequence A--B--C--D--E. The segment distances were used as reference values for evaluating distance estimation, trajectory reconstruction, and final-position error. Therefore, the ground truth in this study represents a checkpoint-based reference trajectory rather than an external GNSS-based trajectory.

For each trial, the recorded trajectory was compared with the checkpoint-based ground truth. The position error at time index $k$ was computed as

\begin{equation}
e_k = \sqrt{(x_k-\hat{x}_k)^2 + (y_k-\hat{y}_k)^2},
\end{equation}

where $(x_k,y_k)$ denotes the ground-truth position and $(\hat{x}_k,\hat{y}_k)$ denotes the estimated position. The overall trajectory accuracy was evaluated using mean error, root mean square error (RMSE), maximum error, and final-position error.

\subsection{Sensor Data Acquisition}

Sensor data are collected from the onboard IMU, magnetometer, and barometer at a fixed sampling rate. The IMU provides tri-axis acceleration and angular velocity measurements used for motion estimation.

IMU calibration is performed to reduce sensor bias and noise before motion estimation. Calibration procedures follow the inertial navigation techniques described in \cite{b7}.

\subsection{Parameter Selection}

The step-detection threshold and adaptive step-length coefficients were selected from the walking trials in the experimental dataset. The acceleration magnitude was first computed from the tri-axis accelerometer signal. The step-detection threshold $T_s$ was then selected based on the mean and standard deviation of the acceleration magnitude during walking:

\begin{equation}
T_s = \mu_{a} + \lambda \sigma_{a},
\end{equation}

where $\mu_{a}$ and $\sigma_{a}$ are the mean and standard deviation of the acceleration magnitude, respectively, and $\lambda$ is a sensitivity coefficient. In this study, $\lambda$ was selected empirically to avoid false step detections caused by small hand or body movements while preserving true walking peaks.

The adaptive step-length model was defined as

\begin{equation}
L_k = \alpha + \beta f_{step,k} + \gamma s_k,
\end{equation}

where $L_k$ is the estimated step length at step $k$, $f_{step,k}$ is the step frequency, and $s_k$ is the slope-related term derived from barometric altitude variation. The coefficients $\alpha$, $\beta$, and $\gamma$ were determined from the experimental trials by fitting the estimated step length to the reference segment distances using least-squares regression:

\begin{equation}
\boldsymbol{\theta} =
\begin{bmatrix}
\alpha & \beta & \gamma
\end{bmatrix}^{T}
=
(\mathbf{X}^{T}\mathbf{X})^{-1}\mathbf{X}^{T}\mathbf{L},
\end{equation}

where $\mathbf{X}$ contains the step-frequency and slope features, and $\mathbf{L}$ contains the reference step-length values obtained from the checkpoint-based route distances. This parameter-selection procedure allows the proposed model to adapt to walking dynamics while maintaining reproducibility.

\subsection{Step Detection Algorithm}

Step detection was performed using the acceleration magnitude computed from the tri-axis accelerometer signal:

\begin{equation}
a_{mag,k} = \sqrt{a_{x,k}^{2}+a_{y,k}^{2}+a_{z,k}^{2}}.
\end{equation}

To reduce high-frequency noise, the acceleration magnitude was smoothed using a moving-average filter:

\begin{equation}
\bar{a}_{mag,k} = \frac{1}{N}\sum_{i=0}^{N-1} a_{mag,k-i},
\end{equation}

where $N$ is the smoothing window length. A valid step was detected when a local peak exceeded the adaptive threshold $T_s$ and the time interval from the previous detected step was greater than a minimum refractory interval $T_{min}$:

\begin{equation}
\bar{a}_{mag,k} > T_s,
\end{equation}

\begin{equation}
\bar{a}_{mag,k} > \bar{a}_{mag,k-1}, \quad
\bar{a}_{mag,k} > \bar{a}_{mag,k+1},
\end{equation}

\begin{equation}
t_k - t_{last} > T_{min}.
\end{equation}

This procedure suppresses false detections caused by transient vibration or small non-walking movements.

\subsection{Heading Estimation Using Gyroscope--Magnetometer Fusion}

Heading estimation was performed by combining short-term gyroscope integration with magnetometer-assisted correction. The gyroscope provides smooth short-term angular motion, while the magnetometer provides an absolute heading reference. To reduce heading drift, a lightweight Extended Kalman Filter (EKF) formulation was used.

The state vector was defined as

\begin{equation}
\mathbf{x}_k =
\begin{bmatrix}
\theta_k \\
b_{\omega,k}
\end{bmatrix},
\end{equation}

where $\theta_k$ is the pedestrian heading angle and $b_{\omega,k}$ is the gyroscope bias. The prediction model was given by

\begin{equation}
\theta_k^{-} = \theta_{k-1} + (\omega_{z,k}-b_{\omega,k-1})\Delta t,
\end{equation}

\begin{equation}
b_{\omega,k}^{-} = b_{\omega,k-1}.
\end{equation}

The magnetometer-based heading measurement was computed as

\begin{equation}
z_k = \operatorname{atan2}(m_{y,k},m_{x,k}),
\end{equation}

where $m_{x,k}$ and $m_{y,k}$ are the horizontal magnetometer components. The EKF update step was then used to correct the predicted heading using the magnetometer measurement. To reduce the effect of magnetic disturbance, the measurement update was weighted according to heading stability. When abnormal magnetic variation was detected, the magnetometer correction was reduced and the system relied more strongly on gyroscope-based heading propagation.

\subsection{Baseline PDR Method}

To provide a fair comparison, a traditional PDR baseline was implemented using constant step-length estimation and direct heading propagation. In the baseline method, the pedestrian position was updated only when a valid step was detected. The position update was computed as

\begin{equation}
\hat{x}_{k} = \hat{x}_{k-1} + L_0 \cos(\hat{\theta}_k),
\end{equation}

\begin{equation}
\hat{y}_{k} = \hat{y}_{k-1} + L_0 \sin(\hat{\theta}_k),
\end{equation}

where $L_0$ is a fixed nominal step length and $\hat{\theta}_k$ is the estimated heading angle. Unlike the proposed ForestBack method, the baseline does not use adaptive step-length estimation, terrain-aware compensation, or breadcrumb-based correction. Therefore, the baseline represents a conventional PDR approach in which trajectory drift accumulates from step-length and heading errors.

The proposed ForestBack method used the same detected step events as the baseline, but replaced the constant step length with the adaptive model and used heading compensation to reduce accumulated drift. This ensured that the comparison focused on the effect of the proposed step-length and heading-stabilization mechanisms.

\subsection{Return Path Navigation Algorithm (Proposed)}

When the user activates return mode, the system reverses the breadcrumb sequence to guide the user back to the starting location.

\begin{equation}
P_n \rightarrow P_{n-1} \rightarrow \dots \rightarrow P_0
\end{equation}

Directional guidance is generated from the stored trajectory.

\subsection{Evaluation Metrics}

The performance of the proposed method was evaluated using trajectory-level and trial-level metrics. The RMSE was computed as

\begin{equation}
RMSE = \sqrt{\frac{1}{N}\sum_{k=1}^{N} e_k^2},
\end{equation}

where $e_k$ is the position error at time index $k$ and $N$ is the number of evaluated samples. The mean position error was computed as

\begin{equation}
E_{mean} = \frac{1}{N}\sum_{k=1}^{N} e_k.
\end{equation}

The final-position error was defined as the Euclidean distance between the estimated final point and the ground-truth final checkpoint:

\begin{equation}
E_{final} = \sqrt{(x_E-\hat{x}_E)^2 + (y_E-\hat{y}_E)^2}.
\end{equation}

Turn-event detection consistency was evaluated by comparing detected turn events with the reference turn labels obtained from the checkpoint route. These metrics were used to compare the proposed ForestBack method with the traditional PDR baseline.

\subsection{Dataset and Reproducibility}

The dataset used in this study is publicly available in the ForestBack-Dataset repository~\cite{b11}. The repository contains the dataset package and analysis notebook used to reproduce the reported results. The dataset includes time-series inertial measurements, magnetometer readings, barometric variables, checkpoint-based ground-truth trajectories, baseline PDR outputs, proposed ForestBack outputs, error metrics, and power-related variables.

The availability of the dataset and notebook allows the trajectory reconstruction results, RMSE analysis, spatial error heatmap, heading-error distribution, and overall quantitative comparison to be independently reproduced. This supports transparent benchmarking of breadcrumb-based PDR methods for infrastructure-free return navigation.

Unlike conventional PDR evaluation that focuses only on endpoint localization error, this study evaluates route-preserving navigation. This is important because return navigation requires the recorded trajectory to be reversible into meaningful guidance commands, not merely close to the final coordinate.

\section{Results and Analysis}

This section presents the experimental results obtained from the proposed ForestBack indoor obstacle-avoidance evaluation. The experiment was designed to assess the performance of the proposed pedestrian dead reckoning (PDR)-based breadcrumb navigation system under a controlled indoor environment containing multiple walking segments, turning points, and a central obstacle. The experimental route was defined using five checkpoints, denoted as A, B, C, D, and E, where checkpoint A was used as the starting point and checkpoint E was used as the final destination. The route was intentionally designed to require obstacle avoidance rather than direct straight-line walking, allowing step detection, heading estimation, trajectory reconstruction, and accumulated drift behavior to be evaluated under a more challenging condition.

A high-resolution experimental dataset was used to evaluate the proposed indoor obstacle-avoidance route. The dataset includes inertial sensor signals, heading estimates, barometric-related variables, step events, turn-event labels, reconstructed trajectories, and error metrics. The dataset consists of 36 walking trials and 42,474 time-series samples collected at an equivalent sampling rate of 100 Hz. Each trial contains the ground-truth checkpoint-based trajectory, the estimated trajectory from a traditional PDR baseline, and the estimated trajectory from the proposed ForestBack method. To support reproducibility and future benchmarking, the complete dataset and analysis notebook have been made publicly available through the ForestBack-Dataset repository~\cite{b11}.

The evaluation was designed not only to measure point-wise localization error but also to examine whether the overall route structure could be preserved. This is important for breadcrumb-based return navigation because the user does not only require an accurate final coordinate; the system must also preserve the sequence of walking and turning actions that can be reversed into practical guidance commands. Therefore, the proposed evaluation emphasizes route reconstruction, error accumulation, heading stability, and final return-path reliability.

\subsection{Trajectory Reconstruction Performance}

\begin{figure}[t]
    \centering
    \includegraphics[width=\linewidth]{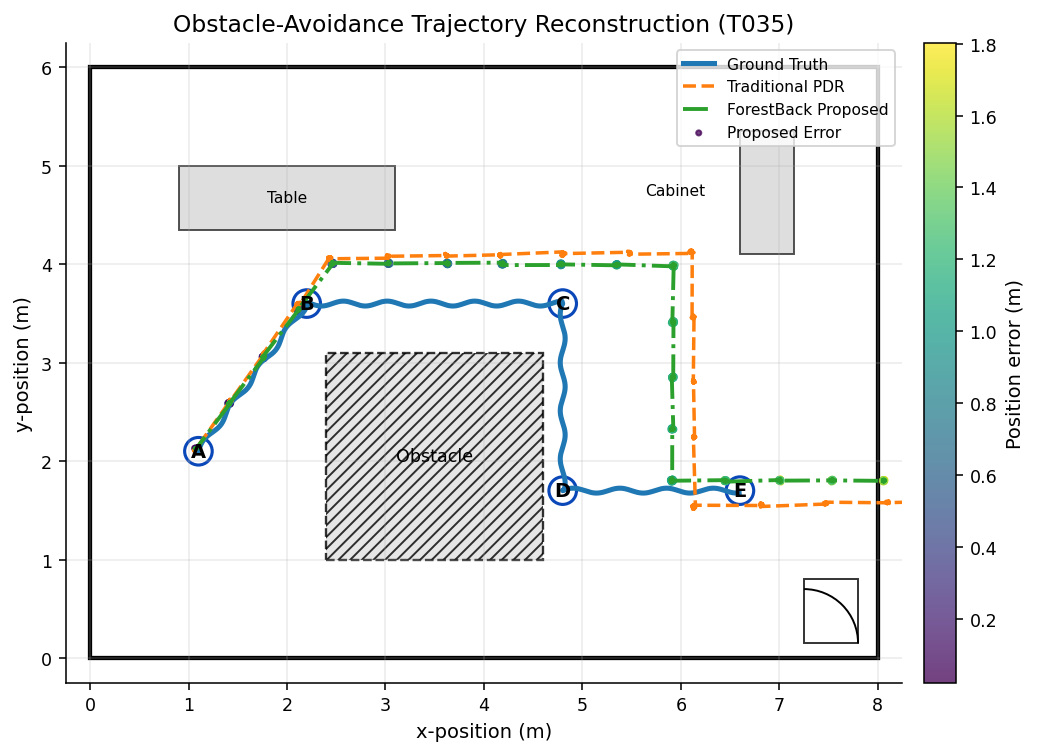}
    \caption{Obstacle-avoidance trajectory reconstruction for a representative trial. The ground-truth route, traditional PDR trajectory, and proposed ForestBack trajectory are compared on the indoor test map. The color-coded points represent the position error of the proposed method.}
    \label{fig:trajectory_reconstruction}
\end{figure}

Figure~\ref{fig:trajectory_reconstruction} shows an example of trajectory reconstruction for the indoor obstacle-avoidance route. The ground-truth path follows the predefined checkpoint sequence A--B--C--D--E, where the participant is required to move around the central obstacle. The traditional PDR trajectory shows a noticeable deviation from the ground-truth route, especially after the first major heading transition. This behavior indicates that accumulated heading error and step-length uncertainty can cause the estimated path to drift away from the intended route.

In contrast, the proposed ForestBack trajectory follows the ground-truth path more closely. The reconstructed path remains near the expected obstacle-avoidance corridor and shows smaller lateral deviation near checkpoints C, D, and E. This improvement can be attributed to the use of adaptive step-length estimation, heading-drift compensation, and breadcrumb-based segment representation. The color-coded error distribution also indicates that the position error increases gradually along the route, but the proposed method is able to reduce the error magnitude compared with the conventional PDR baseline.

\subsection{Spatial Error Distribution}

\begin{figure}[t]
    \centering
    \includegraphics[width=\linewidth]{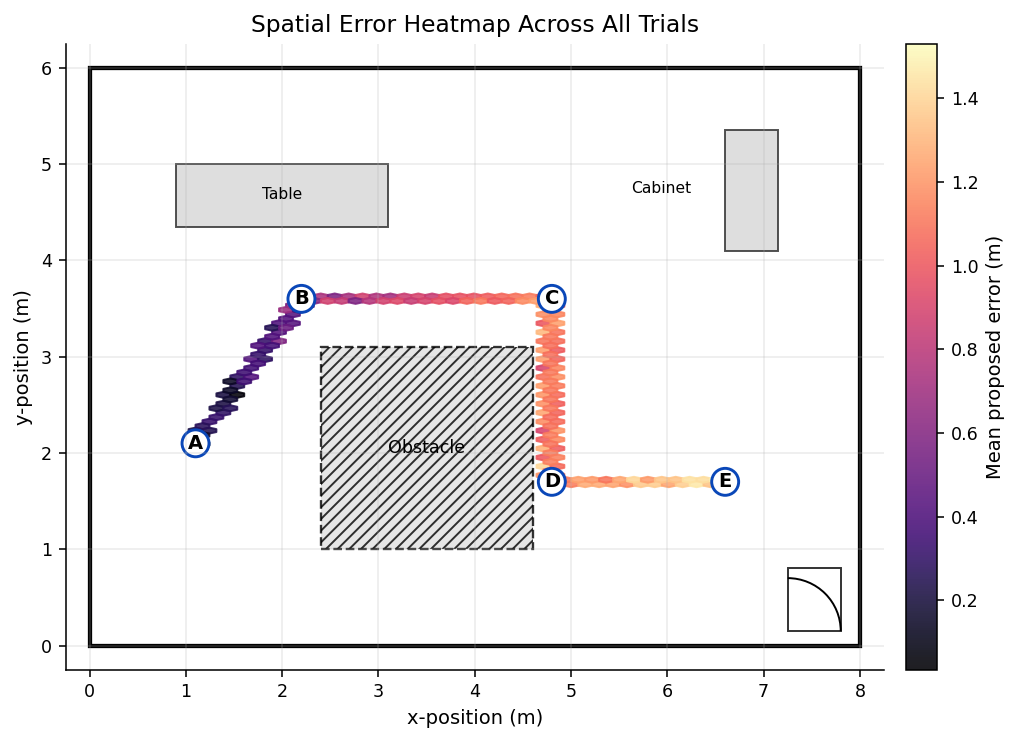}
    \caption{Spatial error heatmap of the proposed ForestBack method across all trials. Higher error regions are mainly concentrated near later route segments and turning regions around the obstacle.}
    \label{fig:spatial_error_heatmap}
\end{figure}

The spatial error heatmap in Fig.~\ref{fig:spatial_error_heatmap} illustrates the mean position error of the proposed ForestBack method across all trials. The error distribution is not uniform along the route. Lower errors are observed near the early route segment from A to B, while higher errors are concentrated near the later route sections around C, D, and E. This behavior is expected because PDR-based navigation accumulates error over time as each new position estimate is computed from the previous estimate.

The highest error regions are observed near the right side of the central obstacle and toward the final checkpoint E. These regions correspond to the parts of the route where multiple heading transitions occur after the user has already traveled a significant distance. Therefore, the spatial heatmap confirms that accumulated heading drift and segment-transition uncertainty are major contributors to final-position error. However, the proposed method maintains the estimated trajectory within the expected obstacle-avoidance corridor, demonstrating that the breadcrumb-based representation can preserve the overall route structure even when local errors increase.

\subsection{Error Accumulation Along the Obstacle-Avoidance Route}

\begin{figure}[t]
    \centering
    \includegraphics[width=\linewidth]{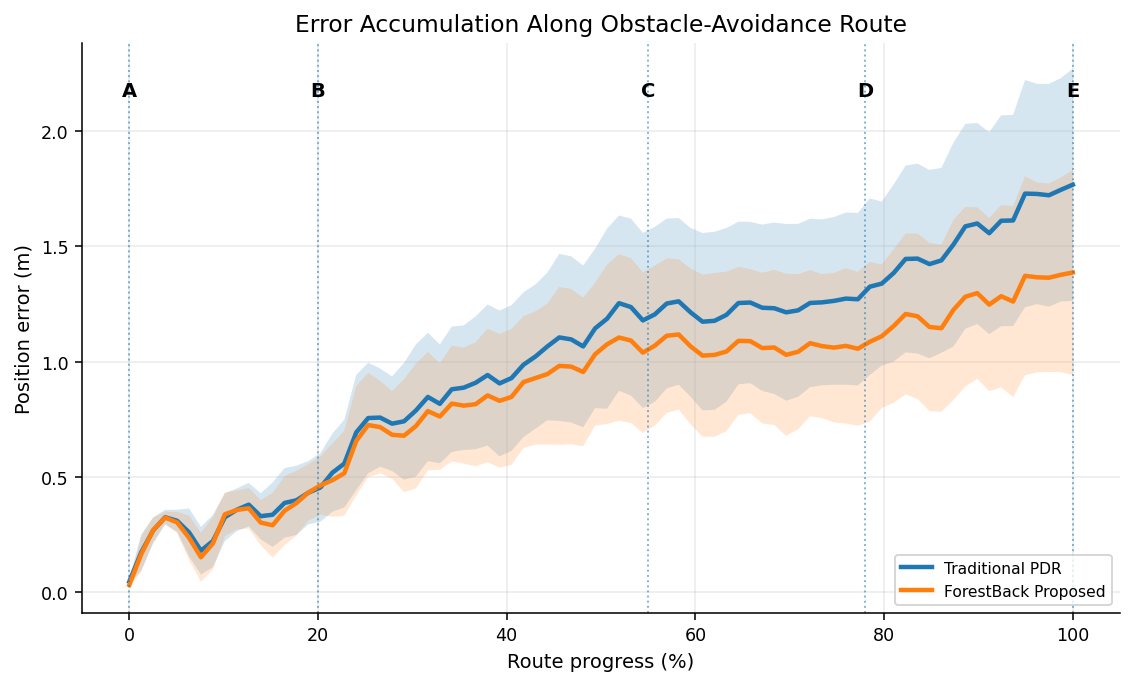}
    \caption{Error accumulation along the obstacle-avoidance route. The proposed ForestBack method shows lower position error than the traditional PDR baseline throughout most of the route, especially after major heading transitions.}
    \label{fig:error_accumulation}
\end{figure}

Figure~\ref{fig:error_accumulation} compares the position error accumulation of the traditional PDR baseline and the proposed ForestBack method along the normalized route progress from checkpoint A to checkpoint E. The vertical dashed lines indicate the approximate checkpoint locations along the route. As expected, both methods show increasing position error as the route progresses because PDR estimation is affected by cumulative step-length and heading errors.

The traditional PDR method exhibits a faster error growth rate, especially after checkpoint B and around the C--D transition. This indicates that the conventional method is more sensitive to heading changes and accumulated drift. In contrast, the proposed ForestBack method produces consistently lower error throughout most of the route. The difference between the two methods becomes more apparent after checkpoint C, where the route requires a vertical transition along the right side of the obstacle. This result suggests that the proposed approach is more robust in handling multi-segment obstacle-avoidance navigation.

At the final checkpoint E, the traditional PDR error reaches approximately 1.7--1.8 m on average, while the proposed ForestBack method remains closer to approximately 1.3--1.4 m. This demonstrates that the proposed method reduces final accumulated error and provides a more reliable route reconstruction for return-path guidance.

\subsection{Trial-Level RMSE Distribution}

\begin{figure}[t]
    \centering
    \includegraphics[width=\linewidth]{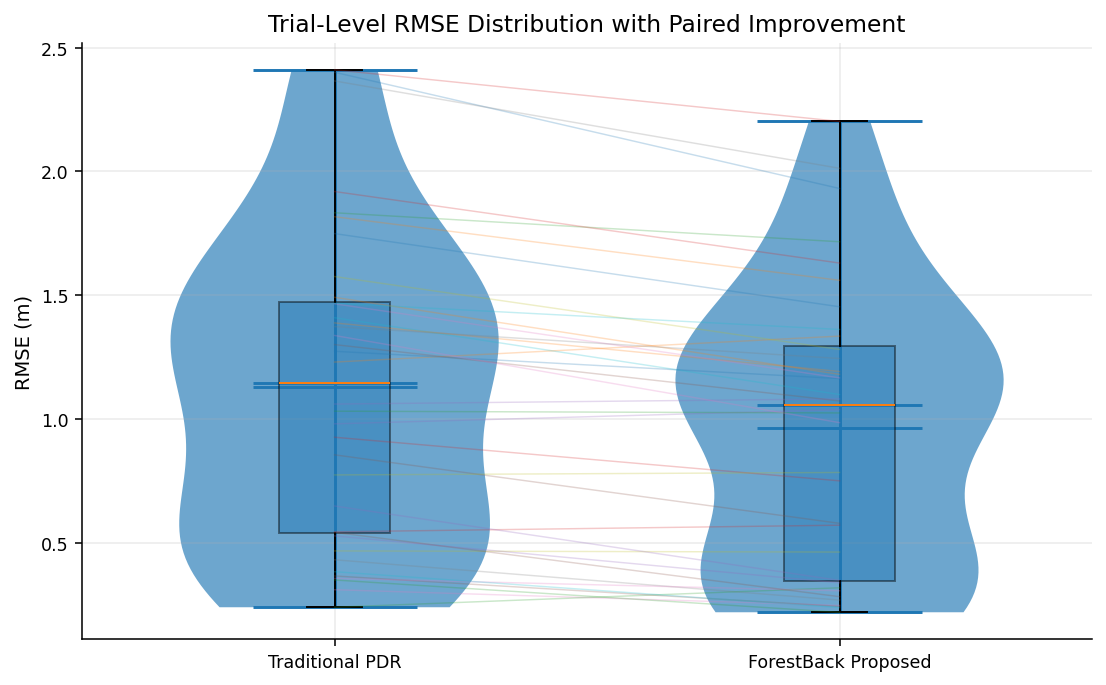}
    \caption{Trial-level RMSE distribution with paired improvement comparison. The proposed ForestBack method achieves lower RMSE and reduced variability compared with the traditional PDR baseline.}
    \label{fig:rmse_distribution}
\end{figure}

The trial-level RMSE distribution is presented in Fig.~\ref{fig:rmse_distribution}. A paired comparison was performed between the traditional PDR baseline and the proposed ForestBack method for each trial. The violin plot shows the distribution shape of RMSE values, while the box plot indicates the median and interquartile range. The connecting lines between the two methods represent paired trial-level improvements.

The traditional PDR method shows a wider RMSE distribution, indicating higher variability across different trials. This suggests that the baseline method is more sensitive to walking variations, sensor noise, mounting conditions, and heading disturbance. In comparison, the proposed ForestBack method produces a narrower distribution and a lower median RMSE. Across all 36 trials, the traditional PDR baseline achieved a mean RMSE of approximately 1.129 m, whereas the proposed ForestBack method achieved a mean RMSE of approximately 0.965 m. This corresponds to an average RMSE improvement of approximately 15.76\%.

The paired improvement lines further show that most trials benefit from the proposed method. Although some trials still contain residual error due to accumulated drift and magnetic disturbance, the overall distribution confirms that the proposed method provides more stable and accurate trajectory reconstruction than the traditional PDR baseline.

\subsection{Heading Error Analysis}

\begin{figure}[t]
    \centering
    \includegraphics[width=\linewidth]{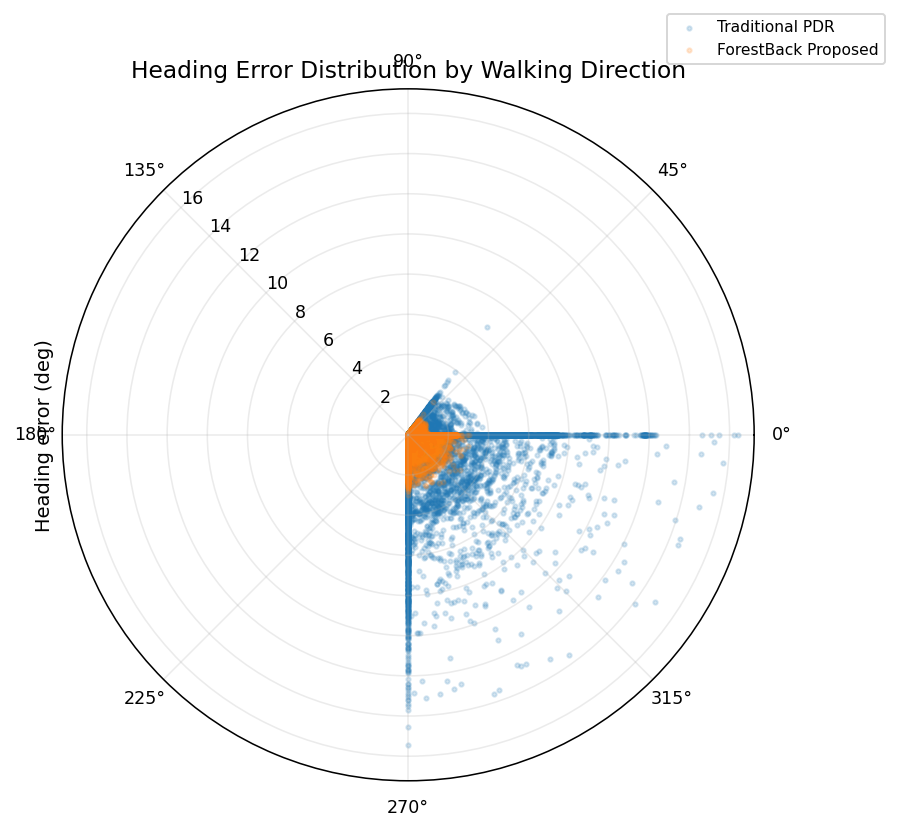}
    \caption{Heading error distribution by walking direction. The proposed ForestBack method shows a more compact error distribution than the traditional PDR baseline, indicating improved heading stability.}
    \label{fig:heading_error_polar}
\end{figure}

Figure~\ref{fig:heading_error_polar} presents the heading error distribution with respect to walking direction. The polar visualization was used to analyze whether the heading error depends on the movement direction of the user. The traditional PDR method produces a wider spread of heading error, particularly along the main horizontal and vertical walking directions. This indicates that heading estimation error can become more significant when the route contains repeated straight segments followed by abrupt turns.

The proposed ForestBack method shows a more compact heading error distribution. The lower radial spread indicates that the proposed heading compensation strategy can suppress drift and reduce the impact of heading uncertainty. Since heading error directly affects PDR trajectory reconstruction, the reduced heading error observed in the proposed method explains the lower RMSE and smaller final-position error reported in the previous sections.

\subsection{Overall Quantitative Performance}

Table~\ref{tab:overall_performance} summarizes the overall quantitative performance of the traditional PDR baseline and the proposed ForestBack method. The results show that the proposed method reduces both RMSE and final-position error. The mean RMSE is reduced from 1.129 m to 0.965 m, while the mean final-position error is reduced from 1.781 m to 1.388 m. 

In addition, the turn-event detection consistency reaches approximately 99.90\%, suggesting that the route structure can be reliably recorded for breadcrumb-based reverse navigation.

\begin{table}[t]
\centering
\caption{Overall performance comparison.}
\label{tab:overall_performance}
\small
\setlength{\tabcolsep}{3pt}
\begin{tabular}{lcc}
\hline
\textbf{Metric} & \textbf{PDR} & \textbf{ForestBack} \\
\hline
RMSE (m) & 1.129 & 0.965 \\
Final error (m) & 1.781 & 1.388 \\
RMSE impr. (\%) & -- & 15.76 \\
Turn consistency (\%) & -- & 99.90 \\
Current (mA) & -- & 60.31 \\
\hline
\end{tabular}
\end{table}

\section{Conclusion}

This paper presented \textit{ForestBack}, an infrastructure-free pedestrian navigation framework for GPS-denied environments. The proposed system combines pedestrian dead reckoning (PDR), inertial sensing, heading estimation, barometric-altitude-assisted correction, and breadcrumb-based return guidance. By recording the walking route as reversible breadcrumb nodes, ForestBack can support return navigation without GPS, Wi-Fi, Bluetooth beacons, or pre-installed infrastructure. The system was evaluated using an indoor obstacle-avoidance route with five checkpoints from A to E. A high-resolution experimental dataset consisting of 36 walking trials and 42,474 time-series samples at 100 Hz was used for analysis. The results showed that ForestBack reduced the mean RMSE from 1.129 m to 0.965 m compared with traditional PDR, corresponding to a 15.76\% improvement. The mean final-position error was also reduced from 1.781 m to 1.388 m, while turn-event detection consistency reached approximately 99.90\%. These results indicate that ForestBack can improve trajectory reconstruction and preserve route structure in obstacle-avoidance navigation scenarios. However, long-term drift remains a limitation because the system relies on relative motion estimation. Future work will focus on real-world hardware validation, outdoor testing, multi-user experiments, and improved drift compensation. The dataset and analysis notebook have been made publicly available through the ForestBack-Dataset repository~\cite{b11} to support reproducibility and future benchmarking.


\begin{thebibliography}{00}
\bibitem{b1} Yuan Wu, Hai-Bing Zhu, Qing-Xiu Du and Shu-Ming Tang. A Survey of the Research Status of Pedestrian Dead Reckoning Systems Based on Inertial Sensors. International Journal of Automation and Computing, vol. 16, no. 1, pp. 65-83, 2019. DOI:  10.1007/s11633-018-1150-y

\bibitem{b2} Kuang J, Niu X, Chen X. Robust Pedestrian Dead Reckoning Based on MEMS-IMU for Smartphones. Sensors (Basel). 2018 May 1;18(5):1391. doi: 10.3390/s18051391. PMID: 29724003; PMCID: PMC5982656.

\bibitem{b3} Geng J, Xia L, Xia J, Li Q, Zhu H, Cai Y. Smartphone-Based Pedestrian Dead Reckoning for 3D Indoor Positioning. Sensors (Basel). 2021 Dec 8;21(24):8180. doi: 10.3390/s21248180. PMID: 34960273; PMCID: PMC8706353.

\bibitem{b4} Wei Li, Ruizhi Chen, Yue Yu, Yuan Wu, Haitao Zhou,
Pedestrian dead reckoning with novel heading estimation under magnetic interference and multiple smartphone postures,
Measurement,
Volume 182,
2021,
109610,
ISSN 0263-2241,
https://doi.org/10.1016/j.measurement.2021.109610.

\bibitem{b5} Zhu P, Yu X, Han Y, Xiao X, Liu Y. Improving Indoor Pedestrian Dead Reckoning for Smartphones under Magnetic Interference Using Deep Learning. Sensors (Basel). 2023 Nov 23;23(23):9348. doi: 10.3390/s23239348. PMID: 38067722; PMCID: PMC10708641.

\bibitem{b6}Kuang, J., Niu, X., Zhang, P., \& Chen, X. (2018). Indoor Positioning Based on Pedestrian Dead Reckoning and Magnetic Field Matching for Smartphones. Sensors, 18(12), 4142. https://doi.org/10.3390/s18124142

\bibitem{b7} F. Montorsi, F. Pancaldi, and G. M. Vitetta. Design and Implementation of an Inertial Navigation System for Pedestrians Based on a Low-Cost MEMS IMU. arXiv preprint arXiv:1503.07889, 2015. Available: https://arxiv.org/abs/1503.07889

\bibitem{b8} M. Osman, F. Viset, and M. Kok, “Indoor SLAM Using a Foot-mounted IMU and the Local Magnetic Field,” arXiv preprint arXiv:2203.15866, 2022. [Online]. Available: https://arxiv.org/abs/2203.15866

\bibitem{b9} Saha SS, Du Y, Sandha SS, Garcia LA, Jawed MK, Srivastava M. Inertial Navigation on Extremely Resource-Constrained Platforms: Methods, Opportunities and Challenges. IEEE ION Position Locat Navig Symp. 2023 Apr;2023:708-723. doi: 10.1109/plans53410.2023.10139997. Epub 2023 Jun 8. PMID: 37736264; PMCID: PMC10512424.

\bibitem{b10} S. Wang and N. S. Ahmad, "A Comprehensive Review on Sensor Fusion Techniques for Localization of a Dynamic Target in GPS-Denied Environments," in IEEE Access, vol. 13, pp. 2252-2285, 2025, doi: 10.1109/ACCESS.2024.3519874.

\bibitem{b11} A. Aueawatthanaphisut, ``ForestBack-Dataset,'' GitHub repository, 2026. [Online]. Available: https://github.com/Aueaphum2541/ForestBack-Dataset

\end{thebibliography}
\end{document}